\documentclass{article}
\usepackage{spconf,amsmath,epsfig}
\usepackage{multirow}
\usepackage{amsthm,amssymb}
\usepackage{booktabs}
 
\usepackage{mathrsfs}
\usepackage{bbding}
\usepackage{adforn}

\newcommand{\MName}{CUS3D}

\let\OLDthebibliography\thebibliography
\renewcommand\thebibliography[1]{
  \OLDthebibliography{#1}
  \setlength{\parskip}{0pt}
  \setlength{\itemsep}{0pt plus 0.3ex}
}

\begin{document}\sloppy

\def\x{{\mathbf x}}
\def\L{{\cal L}}

\title{\MName :~CLIP-based Unsupervised 3D Segmentation via Object-level Denoise}
%


\name{Fuyang Yu$^1$, Runze Tian$^1$, Zhen Wang$^2$, Xiaochuan Wang$^3$, Xiaohui Liang$^{1,*}$}

\address{$^1$ Beihang University, Beijing, China \\
$^2$ Tokyo Institute of Technology, Tokyo, Japan \\
$^3$ Beijing Technology and Business University, Beijing, China\\
$^*$ {\tt liang\_xiaohui@buaa.edu.cn}
}

\maketitle

\begin{abstract}
\noindent To ease the difficulty of acquiring annotation labels in 3D data, a common method is using unsupervised and open-vocabulary semantic segmentation, which leverage 2D CLIP semantic knowledge.
In this paper, unlike previous research that ignores the ``noise'' raised during feature projection from 2D to 3D, we propose a novel distillation learning framework named \MName{}.
In our approach, an object-level denosing projection module is designed to screen out the ``noise'' and ensure more accurate 3D feature.
Based on the obtained features, a multimodal distillation learning module is designed to align the 3D feature with CLIP semantic feature space with object-centered constrains to achieve advanced unsupervised semantic segmentation.
We conduct comprehensive experiments in both unsupervised and open-vocabulary segmentation, and the results consistently showcase the superiority of our model in achieving advanced unsupervised segmentation results and its effectiveness in open-vocabulary segmentation.
\end{abstract}
\begin{keywords}
Unsupervised Semantic Segmentation, Point Cloud, Knowledge Distillation, Multimodal, Open-vocabulary
\end{keywords}
\section{Introduction}
\label{sec:intro}

\begin{figure*}[t]
\centering
\includegraphics[width=0.9\hsize]{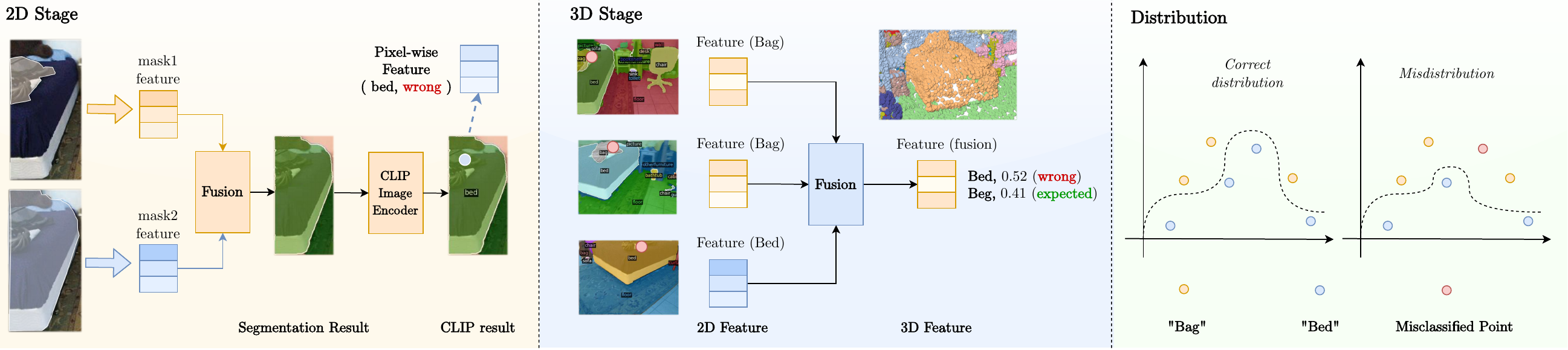}
\caption{
The overview of existing methods (Up) and the analysis of the problems with existing methods (Down).
}
\label{fig:noise}
\end{figure*}
Point cloud semantic segmentation~\cite{zhang2019review,jhaldiyal2023semantic} aiming to give a category label for each input point is a fundamental task of 3D visual understanding.
Due to the challenges in acquiring and annotating point cloud data, there is a shortage of supervised segmentation labels for point clouds, which greatly limit the development of this field and downstream tasks.
To address this shortcoming, many researchers have carried out studies on unsupervised semantic segmentation which means training model without using annotated data.

Several research works have used clustering~\cite{long2023pointclustering,liu2023u3ds} to achieve unsupervised semantic segmentation.
This approach, although applicable to a wide range of scenarios, has limited accuracy and is hard to capture deep semantic context for downstream tasks.
Benefiting from the development of large models such as CLIP~\cite{radford2021learning} which established complete semantic space on 2D image or even biomedical image~\cite{lai2024residualbased}, many researchers~\cite{peng2023openscene,zhang2022pointclip} aimed to achieve unsupervised point cloud semantic segmentation by aligning the 3D semantic space to the 2D semantic space of large models.
The framework of their approaches takes the RGB-D frames as input, and implement 2D semantic segmentation via large model like CLIP.
Pixel-wise features can be obtained through segmented masks, then the features is projected to 3D space using 2D-3D projection methods~\cite{dai20183dmv} to obtain point-wise feature.
The semantic segmentation result can be obtained by calculating the the similarity between the point-wise features and the text embedding of categories.
These approaches leveraging knowledge of large models can achieve more accurate unsupervised semantic segmentation result and have impressive performance on open-vocabulary tasks.

However, though their methods yielded promising results in unsupervised and open-vocabulary semantic segmentation, it is noteworthy that they did not consider the accuracy of the obtained 3D features.
In fact, these 3D features contain a lot of ``noise'', which affects the alignment to the semantic space of large model.
The ``noise'' raised from two aspects.
1) ``Noise'' is introduced by the mask-centered assignment of features in 2D stage.
2) Fusion of features without screening introduces ``noise'' in 3D stage.
In detail, the whole image is divided into patches~\cite{zhang2022pointclip} or masks~\cite{peng2023openscene} for feature exaction in 2D stage.
As shown in Figure \ref{fig:noise} (Yellow), one pixel of the ``bag'' belongs to two different masks.
The big mask is selected as the final results and the CLIP exact its features as ``Bed'' due to the large models pay more attention on the overall semantics.
Masks are difficult to adaptively fit to the size of various objects introducing ``noise''.
In 3D stage, as shown in Figure \ref{fig:noise} (Blue), the red point represent the same point in 3D space and appears in three different frames.
The features projected from the three frames is different, when fusing the feature without screening, the ``noise'' is raised.
Shown in the figure, the point is incorrectly recognized as ``Bad'' finally.

When aligning the semantic space, these ``noises'' affect the accuracy of the learned distributions, as illustrated in Figure \ref{fig:noise} (Green), and further reduce segmentation accuracy.
We visualize the impact of ``noise'' on the segmentation results in Figure \ref{fig:ab&expl} (D).
How to avoid these ``noises'' is a worthy concern and can significantly improve the accuracy of segmentation. 
To address these issues and align the semantic space more accurately, we propose a novel framework \textbf{\MName{}} for unsupervised and open-vocabulary semantic segmentation via object-level denoise.
We believe that pixels or points inside one object should have the same characteristics, the ``noise'' can be suppressed by filtering and assigning features at object-level, but it is difficult to find an accurate object mask without supervision.
To eliminate these noises during projection, we firstly propose a Object-level Denoising Projection(ODP) module, obtaining category to pixels as well as point mask by efficiently clustering and voting strategies, to screen out the ``noise'' raised in 2D stage and 3D stage and obtain more accurate features.
The obtained 3D features are discrete in semantic space, and distillation learning enables the model to learn the complete distribution from discrete sample points and further extends the use of the model.
Thus, we further design a 3D Multimodal Distillation Learning (MDL) module with object-level constraints, constraining the 2D and 3D semantic spaces to be as close as possible centered on the object and further screening out the effects of ``noise''.

In summary, we propose a novel framework called \textbf{\MName{}} to efficiently align the 3D semantic space to the CLIP feature space to achieve advanced unsupervised and open-vocabulary segmentation results, and our contributions in this paper are as follows:
1) We proposed an object-level feature denoise module to ensure more accurate 3D features.
2) We devised a multimodal distillation learning module using object-centered constrains to realize more effectively alignment of the 2D and 3D semantic space.
3) We conducted detailed unsupervised and open-vocabulary segmentation experiments to prove the efficiency of our approach.

\begin{figure*}[t]
\centering
\includegraphics[width=0.9\hsize]{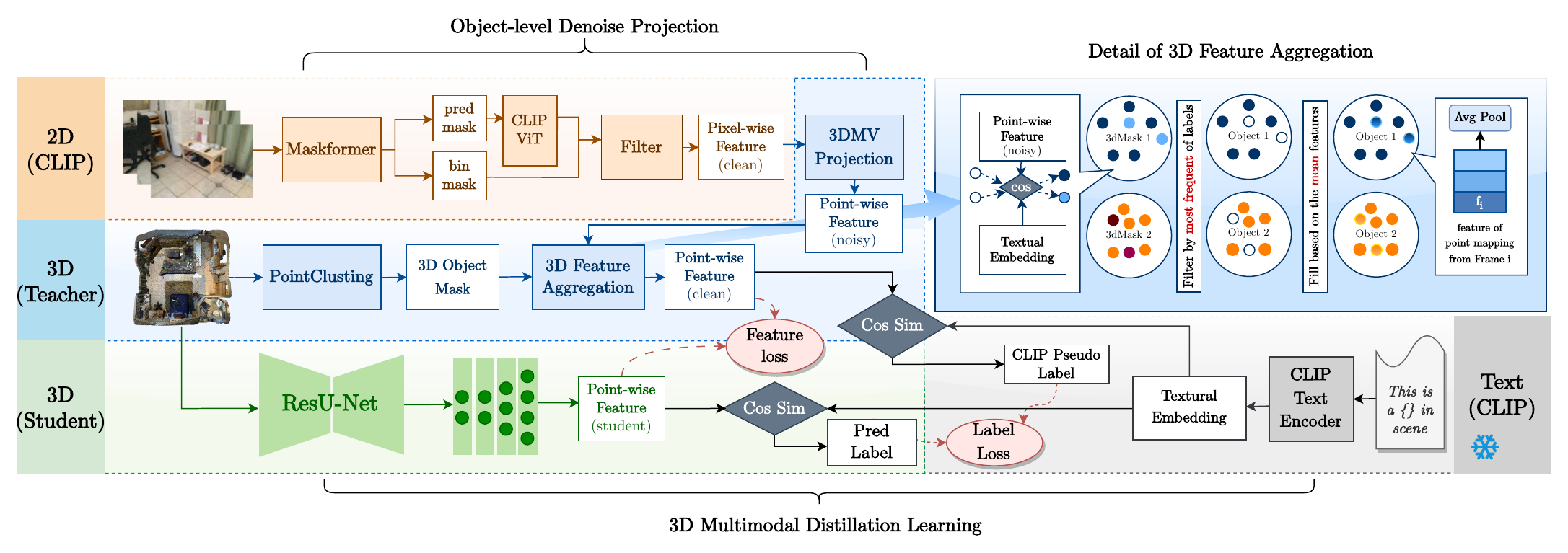}
\caption{
The proposed pipeline comprises two key module, which is described detailly in Section \ref{sec:method}.
}
\label{fig:pipeline}
\end{figure*}

\section{Method}
\label{sec:method}
The overall framework of our proposed method is illustrated in Figure \ref{fig:pipeline}, which consists of four main stages:
1) 2D CLIP feature extraction (orange), extracting pixel-wise features;
2) 3D feature aggregation (blue), obtaining 3D features;
3) 3D student network (green), fitting the semantic space of CLIP;
4) CLIP textual encoder (gray), extracting textual embeddings.
The first and the second stage belongs to the ODP module which denoising at object level to obtain accurate 3D features, while the third and forth stage belongs to the MDL module which fitting the CLIP's semantic space.

\subsection{Object-level Denoising Projection}

This module containing two sub-modules are designed to filter the ``noise'' in the feature at object-level to obtain more accurate 3D features, the two sub-models are described in detail in the following sections.

\subsubsection{2D CLIP Feature Extraction}
\label{sec:2d feature}
We design a 2D feature extraction method to assign pixel-wise features.
Unlike other mask-based methods~\cite{liang2023open}, we assign pixel features centered on the object, and  design a \textbf{2D object-level filter} screening out the ``noise'' raised during feature excation.
We first use a MaskFormer~\cite{cheng2021per} to obtain candidate masks and a pixel-to-mask mapping matrix $M$ ($N \times m$), representing the probability that each pixel belongs to each mask,  
where $N$ is the number of pixels in the image and $m$ is the number of masks. 
Then, for each predicted mask, the corresponding CLIP visual features $F_{CLIP}$ are extracted using the CLIP ViT encoder~\cite{radford2021learning}. 
If a probability value in $M$ is greater than the threshold $\beta$, we assign the corresponding mask feature $F_{CLIP}^{mask_i}$ to this pixel.
Each raw pixel feature is $F_{pixel_j}^{raw} = \left\{ F_{CLIP}^{mask_a},...,F_{CLIP}^{mask_n}\right\}$.
For each raw feature of each pixel, we calculate its category label and obtain a label set of this pixel.
The most frequent category label of the set is voted as the label of this pixel, to establish the connection between pixel and object.
Then we filter out the CLIP feature that does not belong to this category.
Finally, the reserved CLIP features undergo an average pooling to obtain the final pixel-wise feature $F_{pixel}$.

\subsubsection{3D Feature Aggregation}
\label{sec:3d feature}

As shown in the blue part of Figure \ref{fig:pipeline}, the pixel-wise features $F_{pixel}$ obtained from Section \ref{sec:2d feature} are projected to the 3D point clouds using method in 3DMV~\cite{dai20183dmv} to obtain raw point features $F_{point}^{raw}$.
To reduce the ``noise'' in the features, we conduct denoise by a \textbf{3D object-level filter}.
First, we calculate object masks for point clouds,  a pretrained Pointclustering~\cite{long2023pointclustering} is used to obtain the offset of each point cloud to its cluster center, and construct the object mask according to which cluster the point cloud belongs to.
Then, as shown in Figure \ref{fig:pipeline} (3D Feature aggregation), we use the generated object mask to denoise and aggregate the raw point feature.
For each object mask, since the raw point feature is projected from the pixel-wise feature, it also has the CLIP semantic information;
thus, we can measure the cosine similarity between its each raw point-wise feature and the CLIP text embedding, to assign a category label to each raw point-wise feature.
And the most frequent category label among the raw point-wise features is selected as the label for that object mask.
We then screen out any raw point-wise features that do not match this label and each point cloud remains $m$ features.
To get the final point-wise feature $F_{point}$, for each point, 
if $m\geq1$, we perform average pooling on these features to obtain the final feature for that point;
while if $m=0$,
we take the average of all retained features belonging to the same object mask with it and assign this average feature to that point.
This screening process can effectively eliminate noisy features at object level and obtain more robust and accurate point-wise features.

\subsection{3D Multimodal Distillation Learning}
The distribution of 3D features obtained after projection is discrete, in order to align 2D CLIP semantic space more effectively and further eliminate the effects of ``noise'', we adapt a distillation learning network with object-centered constrains to align to the 2D CLIP semantic space.
Using the features (Teacher) obtained in Section \ref{sec:3d feature}, distillation learning can be performed to guide the 3D model to encode the features to fit the CLIP embedding space, thus further suppressing ``noise''.
We supervise the student network (green stage in Figure \ref{fig:pipeline}) with the constrains of features and the labels.
3D ResU-Net~\cite{bhalerao2019brain} is chosen as the backbone network because it can efficiently extract point features and preserve shallow features. 
For a scene with $N$ points, the 3D ResU-Net outputs a feature map of size $N \times 3$, where each of the $N$ points has a corresponding 3D feature vector.
After the backbone network is deployed, we utilize three fully connected layers (MLPs) as projection modules to project the point cloud features into the CLIP embedding space.
Specifically, the output feature map is fed into the projection layers to align the features with the CLIP semantic space. 

\subsubsection{Loss Function}
Relied solely on cosine similarity for supervision can lead to the students network more sensitive to ``noise''.
Unlike previous work, we add object-centered constrain(label loss) for more effectually learning.
The feature loss function maintains semantic consistency between the output features and the target features.
The label loss function provides a soft-supervise signal for training, aligning the two distributions with object-centered constrains which can weak the effect of ``noise'' and learn a rubost distribution.
Using both loss functions together, the model achieves better knowledge distillation performance.

\textbf{Feature Loss} is cosine similarity loss to measure the similarity between target and output features, which is defined as follows: 
\begin{equation}
\small
      L_{feature} =  1 - \frac{1}{N} \sum_{i=0}^N \frac{fo_i \cdot fc_i}{\left|fo_i\right|\left|fc_i\right|}
    \label{formula:feature}
\end{equation}
where $fo_i$ and $fc_i$ represent the features predicted by the network and the aggregated CLIP feature of the $i$-$th$ point, respectively, where $N$ denotes the total number of point clouds.
This loss function constrains the semantic proximity between two features.

\textbf{Label Loss} employs the cross-entropy loss to assess the consistency between the predict results of two features (network output feature and CLIP feature) and is defined as follows:
\begin{equation}
\small
      L_{label} = \frac{1}{N}\sum_{i=0}^N \textit{CrossEntropy}(Pred_i^{output},Pred_i^{CLIP})
    \label{formula:label}
\end{equation}
where $N$ represents the number of point clouds, and $\textit{CrossEntropy}(\cdot,\cdot)$ denotes the cross-entropy loss function. 
$Pred_i^{output}$ and $Pred_i^{CLIP}$ are one-hot matrixes which represent the category prediction results of the two features corresponding to the $i$-$th$ point.
The matrixes are obtained by cosine similarity between the textual embedding and the feature.
Label loss constrains the alignment of the two semantic spaces centered on the object category, enabling the network to further resist the effects of ``noise'' and learn a valid semantic space.
The combined of the two constrains can enable the network to more accurately align CLIP semantic space.

\section{Experiment}
\label{sec:experiment}

\subsection{Dataset and Settings}

\noindent \adfhalfrightarrowhead \, \textbf{Dataset:}
ScanNetV2~\cite{dai2017scannet} and S3DIS~\cite{armeni20163d} are both 3D indoor dataset. ScanNetV2 provide point cloud and RGB-D data of more than 1,500 scenes with objects in 20 categories.
S3DIS provides point cloude and RGB-D data of 6 large areas, 272 rooms, and its object labeled in 13 classes.
 
\noindent \adfhalfrightarrowhead \, \textbf{Implement Details:}
Our experiments were carried out using a GeForce RTX 3090 graphics card with 24GB RAM. 
Our network was distilled using CLIP features in the training set (without labels).
We employ an initial learning rate of 0.001 with cosine annealing learning decay. 
We adapt the accuracy (Acc), mean IoU (mIoU) and harmonic IoU (hIoU) for the evaluation metrics.


\subsection{Quantitative results}

\begin{table}[t]
\begin{tabular}{lcccc}
\toprule
\multirow{2}{*}{\textbf{Model}} & \multicolumn{2}{c}{\textbf{ScanNet}}                         & \multicolumn{2}{c}{\textbf{S3DIS}}     \\ 
 & \multicolumn{1}{c}{\textbf{mIoU}} & \multicolumn{1}{c}{\textbf{Acc}} & \multicolumn{1}{c}{\textbf{mIoU}} & \textbf{Acc} \\ \midrule
\multicolumn{5}{c}{\textbf{Fully-supervised methods}}                                                                          \\ \midrule
 MinkowskiNet~\cite{choy20194d}  &         69.0     &     77.5    & -                      & -  \\
 Mix3D~\cite{nekrasov2021mix3d} &       73.6                &      -                 &    67.2                   &  -\\ 
 Stratified Transformer~\cite{lai2022stratified} &      74.3              &      -                 &    72.0  &  78.1\\ 
 \midrule
 
\multicolumn{5}{c}{\textbf{Unsupervised methods}} \\ 
\midrule
Mseg~\cite{lambert2020mseg} Voting&           45.6            &         54.4      &   42.3   & 51.6  \\
CLIP-FO3D~\cite{zhang2023clip} & 30.2   & 49.1     &    22.3 &  32.8 \\
OpenScene~\cite{peng2023openscene} (LSeg) &      54.2       &    66.6     &   -                    & - \\ 
OpenScene~\cite{peng2023openscene} (OpenSeg) &      47.5       &   70.7     &  -                     &  -\\ 
\midrule
\textbf{\MName{} (Ours)} &      \textbf{57.4} &   \textbf{75.9} & \textbf{52.6}& \textbf{72.6} \\ 
 
\textit{w/o pretraining} &  - &  - & 25.6 & 50.6 \\ 
 \bottomrule
\end{tabular}

\caption{segmentation results on ScanNetV2~\cite{dai2017scannet} (Validation Set) and S3DIS~\cite{armeni2017joint}, our method achieves SOTA unsupervised segmentation performance. 
}
\label{tab:quantitative}
\end{table}

\begin{figure*}[th]
    \centering
    \includegraphics[width=0.932\hsize]{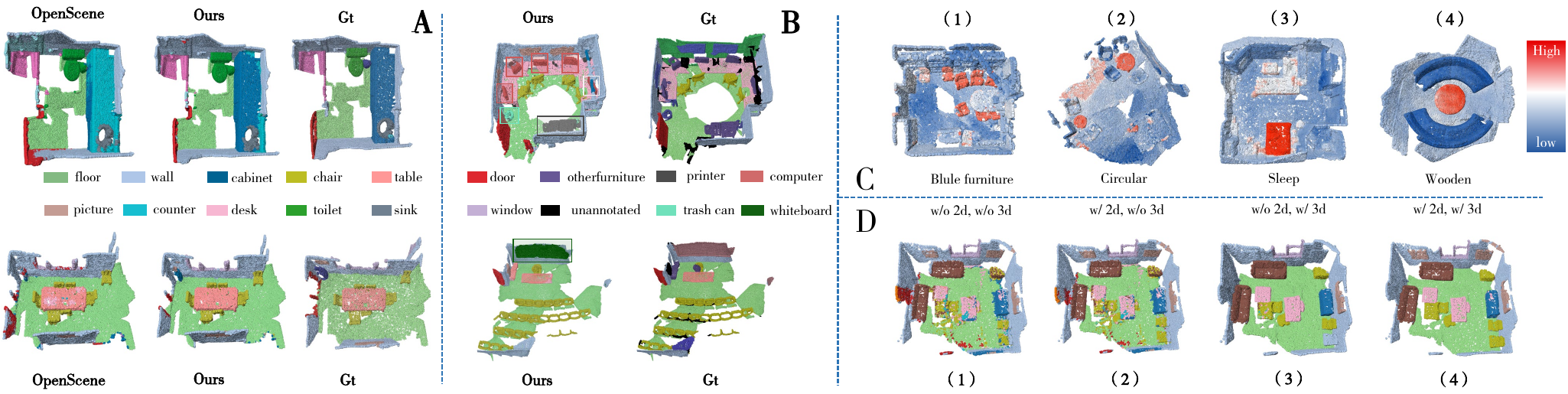}
    \caption{Some Semantic segmentation visualization results.
    }
    \label{fig:ab&expl}
\end{figure*}

\begin{table}[t]
\centering
\small
\begin{tabular}{l|ccc|ccc}
\hline
\multicolumn{1}{c|}{Setting}   & \multicolumn{3}{c|}{Unseen-6}      & \multicolumn{3}{c}{Unseen-10}                     \\ \hline
\multicolumn{1}{c|}{\multirow{2}{*}{Metric}} & \multicolumn{2}{c}{mIoU} & \multirow{2}{*}{hIoU} & \multicolumn{2}{c}{mIoU} & \multirow{2}{*}{hIoU} \\
\multicolumn{1}{c|}{}    &           $\mathcal{S}$  &  $\mathcal{U}$   &     &     $\mathcal{S}$  &  $\mathcal{U}$    &                   \\ \hline
3DGenZ~\cite{michele2021generative}   &   31.2    &  4.8   &   8.3    &   30.1     &  1.4    & 2.7      \\
TGP~\cite{chen2022zero} &   55.2        &   15.4       &   24.1      &     52.5      &    9.5      &      16.1    \\
CLIP-FO3D~\cite{zhang2023clip} &    67.3       & 50.8       &     57.9   &     67.7      &     40.7     &    50.8    \\
\MName{} (Ours)               &    \textbf{68.3}      &     \textbf{53.2}    &    \textbf{59.8}    &    \textbf{69.4}      &   \textbf{46.2}& \textbf{55.5}     \\ \hline
\end{tabular}
\caption{Open-vocabulary semantic segmentation on ScanNetV2. 
``Unseen-i'' indicates that there are i classes that do not have labels during training.
$\mathcal{S}$ and $\mathcal{U}$ represent the performance of seen and unseen classes, respectively.
}
\label{tab:unseen}
\end{table}

\noindent  \adfhalfrightarrowhead \, \textbf{Unsupervised Semantic Segmentation:}
We conducted unsupervised segmentation experiments on both ScanNetV2 and S3DIS. 
In both the feature projection process and the distillation study process, no labels were used. 
The model was subsequently tested on ScannetV2's validation set and S3DIS. 
The experimental results are detailed in Table \ref{tab:quantitative}.
While our method may not surpass existing supervised methods, it has attained state-of-the-art (SOTA) performance among unsupervised approaches on ScanNetV2 (57.4\% vs. 54.2\%).
Notably, we show a huge improvement in accuracy over the other methods on ScanNetV2 (75.9\% vs. 70.7\%), which we attribute to pseudo label produced by the ODP module to guide the network to learn the boundaries of the object and improve the accuracy of semantic segmentation.

\noindent  \adfhalfrightarrowhead \, \textbf{Open-vocabulary Semantic Segmentation:}
We conducted open-vocabulary experiments in two ways on ScanNetV2 dataset. 
In the first experiment, following the settings of CLIP-FO3D~\cite{zhang2023clip}, we split the labels into visible part and invisible part. We only train the model using visible part and then test the model on both visible and invisible labels, to see if our model can segment invisible objects. 
The invisible labels are set to 6 and 10, respectively. 
The results of these experiments are presented in Table \ref{tab:unseen}, 
and our method achieves SOTA in both mIoU and hIoU on Unseen-6 and Unseen-10,
proving its robust open-vocabulary capability.
In the second experiment, as shown in Table \ref{tab:quantitative} (w/o pretraining), we distillate our model on ScanNetV2 training set with 20 labels, and test our model on S3DIS with unseen labels and scenes. It is noticed that our model performs better than CLIP-FO3D (25.6\% vs. 22.3\%) in same settings.
We believe that our two modules weak the effect of the ``noise'' in feature and achieved more effective and robust alignment from 3D feature to 2D CLIP semantic space, thus performing better on onpen-vocabulary semantic segmentation.

\subsection{Ablation Studies}

\begin{table}[t]
\begin{tabular}{cc|cc|cc}
\hline
\multicolumn{2}{c|}{Experiment Settings} & \multicolumn{2}{c|}{S3DIS}    & \multicolumn{2}{c}{ScanNetV2}     \\ \hline
\multicolumn{1}{c|}{Feature Loss} & Label Loss & \multicolumn{1}{c|}{mIoU} & Acc & \multicolumn{1}{c|}{mIoU} & Acc  \\ \hline
\multicolumn{1}{c|}{\Checkmark} & \XSolidBrush & \multicolumn{1}{c|}{49.8} &70.1  & \multicolumn{1}{c|}{53.3} & 74.7 \\ 
\multicolumn{1}{c|}{\XSolidBrush} & \Checkmark  & \multicolumn{1}{c|}{9.2} & 54.3 & \multicolumn{1}{c|}{8.0} & 58.4 \\ 
\multicolumn{1}{c|}{\Checkmark} & \Checkmark  & \multicolumn{1}{c|}{\textbf{52.6}} & \textbf{72.6} & \multicolumn{1}{c|}{\textbf{57.4}} & \textbf{75.9} \\ \hline
\end{tabular}

\caption{Ablation experimental on loss functions.
}
\label{tab:ablation_loss}
\end{table}

\begin{table}[t]
\centering
\begin{tabular}{ccc|cc|cc}
\toprule
\multicolumn{3}{c|}{Experiment Settings}& \multicolumn{2}{c|}{Training Set}    & \multicolumn{2}{c}{Validation Set}  \\ \hline
\multicolumn{1}{c|}{2D FE.} & \multicolumn{1}{c|}{3D FA.} &Sn. & \multicolumn{1}{c|}{mIoU} & Acc & \multicolumn{1}{c|}{mIoU} & Acc \\ \hline

\XSolidBrush   &  \XSolidBrush  &  \XSolidBrush &  27.6  & 47.7 &  28.2 & 45.9 \\
\Checkmark   &  \XSolidBrush  &  \XSolidBrush &   32.4  & 53.2  &   31.8 & 53.0 \\
\XSolidBrush  & \Checkmark    &  \XSolidBrush &  46.7   &  65.3  &  47.3  & 66.8 \\
\Checkmark  & \Checkmark    & \XSolidBrush &   52.9   & 75.5 &   52.7 &  73.9\\
\hline
\XSolidBrush   &  \XSolidBrush  &  \Checkmark &  36.6 & 52.8 &  31.3  & 48.8 \\
\Checkmark   &  \XSolidBrush  &  \Checkmark &  40.3  & 62.3 & 38.6  & 59.5 \\
\XSolidBrush  & \Checkmark    &  \Checkmark & 56.3 & 75.8 &  51.4  &  71.6 \\
\Checkmark  & \Checkmark    &  \Checkmark &  63.8 &  79.4  & 57.4   & 75.9  \\
 \bottomrule
\end{tabular}
\caption{Ablation experiments on ODP module.
In the table, ``2D FE.'' and ``3D FA.'' indicate the usage of 2D feature extraction and  3D feature aggregation module, separately.
``Sn.'' indicates whether using student network in our model.
}
\label{tab:ablation_feature}
\end{table}

In this section, we conduct ablation experiments on the S3DIS and ScanNetV2 datasets, projecting CLIP features for unsupvised segmentation networks distillation, and making segmentation predictions.

We first determine whether our loss function improves the model performance on two datasets.
As shown in Table \ref{tab:ablation_loss}, the network encounters performance drop when only using label loss for distillation study,
while without label loss and only retain feature loss, the performance of the model also drops. 
Using both label loss and feature loss can achieve more accurate segmentation result on both S3DIS dataset (52.6\% vs. 49.8\%) and ScanNetV2 dataset (57.4\% vs. 53.3\%).
The label loss force the two distributions close centered on the object-level, with little attention to detail, and is therefore poorly effective on its own.
The feature loss can bring the two distributions as close as possible, but is susceptible to ``noise''.
When the two loss functions are used together, the network is able to effectively eliminate ``noise'' interference and learn a more robust distribution.


Then to investigate the effectiveness of ODP module, we conduct ablation experiments on the sub-modules in ODP on ScanNetV2 dataset both on training set and validation set.
As indicated in Table \ref{tab:ablation_feature}, 
the four rows above show the mIoU and accuracy calculating bwteen ground truth and pseudo CLIP feature, while the following four rows show the mIoU and accuracy calculate between ground truth and student network's outputs.
It can be found that using student network achieves better semantic segmentation results than directly leveraging pseudo CLIP features (57.4\% vs. 52.7\%) due to that the student network can complement discrete distributions and further resist the effects of ``noise''.

Then we investigate the efficiency of 2D feature extraction and 3D feature aggregation.
Using either 2D feature extraction or 3D feature aggregation module can eliminate the effects of ``noise'' and improve the performance of the models, while combining both of them can achieve even greater improvement on training and validation set, regardless of whether the student network is used.
And compared with 2D feature extraction, 3D feature aggregation has more impact on model performance.


\subsection{Segmentation Visualization}

Figure \ref{fig:ab&expl} gives the visualization results of some experiments.
\textbf{Subfigure A} shows the unsupervised semantic segmentation results on ScannetV2 comparing our method to OpenScene~\cite{peng2023openscene}.
It can be seen that our model perform better in detail of different objects.
\textbf{Subfigure B} demonstrates our model's abilities in open-vocabulary segmentation.
Compared to ground truth, our model can segment objects in unseen categories, such as computers, blackboards, etc.
In \textbf{Subfigure C}, 
using the text below the images, our model can correctly focus on corresponding objects, and the attention results are visualized by heat maps.
This proves that our model is capable of exploring open-vocabulary 3D scenes by not only categories but also other object properties, such as colors, shapes, usages, materials, and so on.
\textbf{Subfigure D} shows the visualization results of the ablation experiments performed on the ODP module. 
This figure clearly visualizes the effect of ``noise'' on the segmentation results, and Both the two sub-modules in ODP can weak the effect of the ``noise'', thus improving the accuracy of the 3D features, while using them both can achieve significantly better results.


\section{Conclution}
\label{sec:conclution}

This paper proposes \MName{} aiming to align 3D features and 2D CLIP semantic space effectually and realise advanced open-vocabulary semantic segmentation. 
Our experimental results consistently demonstrate that our approach outperforms in achieving superior unsupervised segmentation results and exhibits robust open world segmentation capabilities.
Small objects pose a challenge for our segmentation process, often resulting in inaccurate segmentation. 
This occurs because the 2D feature extraction process overlooks the unique characteristics of small objects. 
We plan to address this limitation in future work.

\bibliographystyle{IEEEbib}
\bibliography{icme2023template}

\end{document}